\documentclass[letterpaper]{article} 
\usepackage{aaai2026}  
\usepackage{times}  
\usepackage{helvet}  
\usepackage{courier}  
\usepackage[hyphens]{url}  
\usepackage{graphicx} 
\usepackage{natbib}  
\usepackage{caption} 
\usepackage{algorithm}
\usepackage{algorithmic}
\usepackage{amsmath}
\usepackage{amsfonts}
\usepackage{booktabs}
\usepackage{multirow}
\usepackage{makecell}
\usepackage{newfloat}
\usepackage{listings}
\DeclareCaptionStyle{ruled}{labelfont=normalfont,labelsep=colon,strut=off} 
\floatstyle{ruled}
\newfloat{listing}{tb}{lst}{}
\floatname{listing}{Listing}
\title{ProtTeX-CC: Activating In-Context Learning in Protein LLM via Two-Stage Instruction Compression}
\author{
    Chuanliu Fan\textsuperscript{\rm 1}\equalcontrib,
    Zicheng Ma\textsuperscript{\rm 3,6}\equalcontrib,
    Jun Gao\textsuperscript{\rm 4},
    Nan Yu\textsuperscript{\rm 1}, \\
    Jun Zhang\textsuperscript{\rm 3},
    Ziqiang Cao\textsuperscript{\rm 1,2},
    Yiqin Gao\textsuperscript{\rm 5},
    Guohong Fu\textsuperscript{\rm 1,2}
}

\affiliations{
    \textsuperscript{\rm 1} School of Computer Science and Technology, Soochow University, Suzhou, China\\
    \textsuperscript{\rm 2} Institute of Artificial Intelligence, Soochow University, Suzhou, China\\
    \textsuperscript{\rm 3} Changping Laboratory, Beijing, China\\
    \textsuperscript{\rm 4} Zhejiang University, Hangzhou, China \\
    \textsuperscript{\rm 5} Beijing National Laboratory for Molecular Sciences, College of Chemistry and Molecular Engineering, Peking University, Beijing, China\\
    \textsuperscript{\rm 6} Academy for Advanced Interdisciplinary Studies, Peking University, Beijing, China\\
    20234027004@stu.suda.edu.cn
}

\usepackage{bibentry}
\nocopyright
\begin{document}

\maketitle

\begin{abstract}
Recent advances in protein large language models, such as ProtTeX, represent both side-chain amino acids and backbone structure as discrete token sequences of residue length.
While this design enables unified modeling of multimodal protein information, it suffers from two major limitations: 
(1) The concatenation of sequence and structure tokens approximately doubles the protein length and breaks the intrinsic residue-level alignment between modalities.
(2) Constrained by the training corpus and limited context window, ProtTeX is typically trained on single-protein inputs, rendering it incompatible with in-context learning (ICL) and thus limiting its generalization capability.
To address these issues, we propose ProtTeX-CC, a lightweight two-stage compression framework designed to enhance ProtTeX under few-shot settings.
We first design a joint embedding compression mechanism that fuses sequence and structure representations at the residue level, effectively reducing the protein input length by half without sacrificing performance.
Then we propose a self-compression module that aggregates each full demonstration into the latent space of the last few linguistic tokens, reducing the average demonstration length from $751$ tokens to less than $16$ tokens.
Compared to the original ProtTeX, our self-compression approach achieves a compression ratio of approximately $93.68\%$ in the total prompt length under the 16-shot setting.
Without modifying the backbone model, ProtTeX-CC introduces only a small number of additional parameters through PEFT-based tuning in the joint embedding compression stage and a single trainable projection layer in the self-compression stage.
Extensive experiments on protein function prediction show that ProtTeX-CC improves performance on the in-domain benchmark by $2\%$, and generalizes well to the out-of-domain dataset with a performance gain of $11\%$.
\end{abstract}

\section{Introduction}

\begin{figure}[t]
\centering
\includegraphics[width=0.40\textwidth]{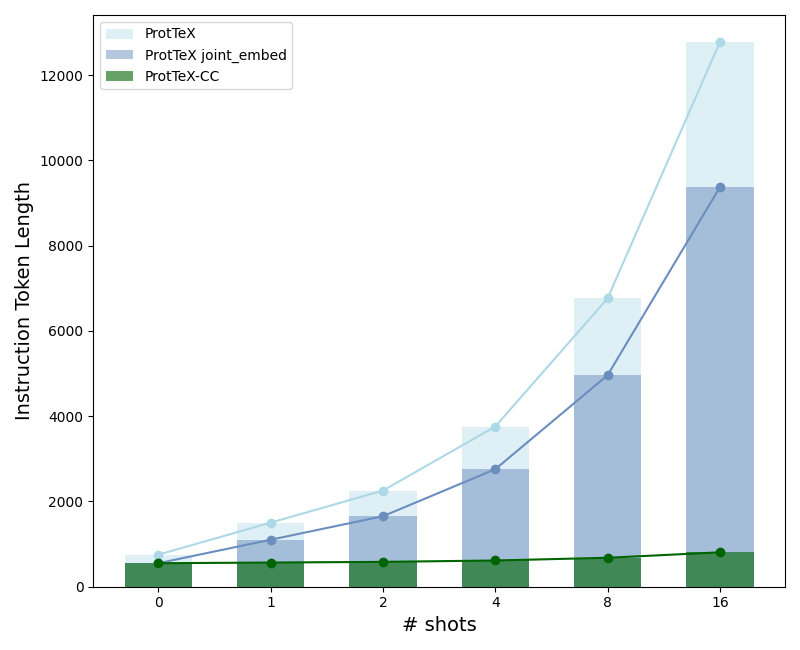}
\caption{Illustration of instruction token lengths for the original ProtTeX, $\text{ProtTeX}_{\text{joint embed}}$, and ProtTeX-CC under few-shot configurations.}
\label{fig_memory}
\end{figure}

Understanding protein function is a fundamental task in computational biology, and recent efforts have explored leveraging large language models (LLMs) to solve the protein question answering (QA) task~\cite{luo2023biomedgptopenmultimodalgenerative,wangProtChatGPTUnderstandingProteins2025}. 
A variety of protein large language models have shown promise in addressing this problem, with models such as ProtT3~\cite{liu2024prott3proteintotextgenerationtextbased}, InstructPLM\cite{Qiu2024.04.17.589642}, and EVOLLAMA~\cite{liu2024evollamaenhancingllmsunderstanding} differing in how they integrate amino acid sequence and backbone structural features.
Among them, recent advances like ProtTeX~\cite{ma2025prottex} have achieved state-of-the-art performance by discretizing both side-chain amino acids and backbone structure into tokens of amino acid residue length.
However, this design introduces two critical bottlenecks.
On the one hand, separating protein sequence and structure tokens nearly doubles the protein length, and disrupts the intrinsic residue-level correspondence between side-chain amino acids and backbone structure.
On the other hand, in-context learning (ICL) is an emergent capability of large language models, allowing them to perform downstream tasks without parameter updates by conditioning on a few task demonstrations provided in the prompt~\cite{brown2020language, bommasani2021opportunities}.
However, ProtTeX is typically trained on single-protein inputs due to inherent constraints in context window length and the scarcity of multi-protein training data. 
This results in a pronounced performance drop when applying naive in-context learning as adding demonstrations may introduce noise rather than help, limiting their few-shot generalization capabilities across new protein QA tasks. Moreover, a critical challenge in protein function prediction is the significant drop in accuracy when models are applied to proteins with low sequence similarity to the training data—i.e., Out-of-Domain (OOD) proteins—where performance can deteriorate to levels close to random guessing~\cite{doi:10.1126/science.adq2634, 10.1093/bib/bbae050}. Given the vast size and diversity of the protein sequence space, harnessing in-context learning (ICL) to retrieve and incorporate high-similarity proteins as contextual examples—without requiring additional training on OOD data—could substantially improve model robustness and accuracy. This approach offers a promising pathway toward enhancing generalization to novel protein queries.

To address these challenges, we propose ProtTeX-CC, a lightweight compression framework that activates in-context learning in ProtTeX through a two-stage compression strategy.
We first fuse the protein sequence and structure embeddings into residue-level joint representations, effectively reducing the protein token length by half while enhancing cross-modal alignment. 
Next, our early experiments revealed that ProtTeX places significantly more attention on the final linguistic modality than on the protein modality, as shown in Table~\ref{tab_atten}.
Motivated by this, we project entire protein QA demonstrations into the latent space of the last few linguistic tokens, enabling compact and semantically meaningful representation of demonstrations within constrained context budgets as illustrated in Figure~\ref{fig_memory}.
Furthermore, the compressed tokens support efficient similarity-based retrieval for demonstration selection. 
This facilitates the construction of dynamic few-shot contexts by retrieving semantically relevant examples based on embedding similarity, rather than relying on lexical matching.

Notably, ProtTeX-CC does not modify the architecture of the backbone ProtTeX model and introduces only a lightweight number of additional parameters through PEFT-based tuning in the joint embedding compression stage and a single trainable projection layer in the self-compression stage.
This design ensures minimal overhead while significantly enhancing the model’s few-shot reasoning ability under realistic context length constraints.

Extensive experiments on protein function prediction benchmarks PFUD~\cite{ma2025prottex} and UniProtQA~\cite{luo2023biomedgptopenmultimodalgenerative} demonstrate the effectiveness and generalization capability of our approach.
ProtTeX-CC is trained solely on the PFUD dataset, and evaluated on both PFUD (in-domain) and UniProtQA (out-of-domain). It achieves state-of-the-art performance on PFUD and exhibits strong generalization to the unseen UniProtQA benchmark, with up to a $11\%$ EMJI improvement over the ProtTeX baseline.

Our main contributions are as follows:
\begin{itemize}
\item To our knowledge, we present the first protein large language model that effectively supports in-context learning, significantly enhancing generalization capability.
\item We propose joint embedding compression, a residue-level compression method that substantially reduces input length while preserving comparable performance to the original model.
\item We further explore the self-compression algorithm, demonstrating that the last few tokens can aggregate contextual information and serve as compressed demonstrations for in-context learning.
\end{itemize}

\section{Related Work}
\subsection{In-Context Learning}
In-context learning (ICL) refers to the phenomenon in which large generative pretrained transformers perform tasks without parameter updates, by conditioning on a few labeled examples provided in the input context~\cite{brown2020language,bommasani2021opportunities}.
Recent advancements have inspired extensions to the multimodal domain~\cite{yang2024exploringdiverseincontextconfigurations,zhao2024mmiclempoweringvisionlanguagemodel}.
One prominent line of work incorporates multi-modal in-context learning directly into the pre-training phase. This typically involves constructing training samples with interleaved multi-modal content (e.g., image-text or protein-text pairs) using handcrafted or templated prompts~\cite{alayrac2022flamingovisuallanguagemodel,awadalla2023openflamingo}. 
However, recent studies on multimodal ICL overlook deployment constraints to some extent~\cite{gao2025aim}, as inference on long contexts of inputs is also time and memory-expensive, as illustrated in Figure~\ref{fig_memory}.
This challenge is especially pronounced when deploying open-source models with limited context lengths, making practical few-shot ICL difficult to achieve.

In contrast to the emergent in-context capabilities exhibited by pre-trained LLMs, models that do not incorporate ICL during pre-training phase have to learn to perform in-context learning explicitly.
AIM~\cite{gao2025aim} introduces a lightweight framework that aggregates vision-language demonstration information into textual labels, while SelfCP~\cite{gao2024selfcp} leverages the target LLM itself to compress over-length textual prompts into dense latent vectors.
Building upon these ideas, our approach further compresses the protein modality into fixed-length token representations, allowing efficient in-context learning without exceeding context limits.
To the best of our knowledge, this work is the first to enable few-shot in-context learning in protein LLM, thus improving generalization ability across protein function understanding tasks.

\subsection{Protein Large Language Models}
The emergence of protein LLMs offers a promising direction for rethinking classical protein function prediction tasks~\cite{abdineProt2TextMultimodalProteins2024a,lvProLLaMAProteinLanguage2024,luoBioMedGPTOpenMultimodal2023,wangProtChatGPTUnderstandingProteins2025}.
These models leverage pretrained large language models to generate structured biological insights about proteins, often in a generative or instruction-following paradigm.
Existing protein LLMs can be broadly categorized into four classes based on their integration strategy for protein inputs: 
(1) Q-Former-based model (e.g., ProtT3~\cite{liu2024prott3proteintotextgenerationtextbased}), which follows the paradigm of InstructBLIP~\cite{dai2023instructblipgeneralpurposevisionlanguagemodels} from the vision-language modeling domain by encoding protein sequence features into fixed-length embeddings using a set of learnable queries; 
(2) Cross-attention-based model (e.g., InstructPLM~\cite{Qiu2024.04.17.589642}), which resembles Qwen-VL~\cite{bai2023qwenvlversatilevisionlanguagemodel} in employing a cross-attention module between protein embeddings and learnable query tokens;  
(3) Projection-based model (e.g., EVOLLAMA~\cite{liu2024evollamaenhancingllmsunderstanding}), which resembles LLaVA-1.5~\cite{liu2024improvedbaselinesvisualinstruction} by utilizing separate encoders for sequence and structure modalities before projecting them into the LLM space;
(4) Discrete token-based models, most notably ProtTeX~\cite{ma2025prottex}, which follow the Chameleon~\cite{chameleonteam2025chameleonmixedmodalearlyfusionfoundation} paradigm by representing both side-chain sequences and backbone structures as discrete token sequences of amino acid residue length.
Among these, ProtTeX has demonstrated state-of-the-art performance on various protein-related tasks, primarily attributed to the mixed-modal early fusion strategy and its powerful conformer encoder ProTokens~\cite{lin2025unifying,Lin2023.11.27.568722}.

However, unlike models that constrain protein information into fixed-length token sequences using learnable query embeddings (e.g., via Q-Former or cross-attention), ProtTeX doubles the input length by concatenating sequence and structure tokens, each spanning the full residue length.
To be specific, the mixed-modal early fusion strategy results in hundreds of tokens per protein input, creating severe context-length bottlenecks.
Therefore, despite its strong performance, ProtTeX suffers from key limitations when applied to in-context learning, primarily due to its doubled protein length and lack of explicit interleaved cross-modal pretraining.
Our work builds upon ProtTeX and directly addresses these context-related limitations through a two-stage compression framework, enabling protein LLMs to perform efficient few-shot in-context learning for the first time.

\section{Methodology}
\subsection{Preliminary}
\label{subsec_preliminary}
Protein LLMs formulate protein function understanding as a text generation task, where the model is prompted with a natural language query and a protein input, and expected to generate a corresponding answer. 
In the standard protein QA setting of ProtTeX, the input follows an interleaved multimodal format composed of the protein's amino acid sequence tokens $t_{\mathbf{s}}=(t_{\mathbf{s},1},t_{\mathbf{s},2},\dots,t_{\mathbf{s},\text{N}_{\text{res}}})$ and the structure tokens $t_{\mathbf{x}}=(t_{\mathbf{x},1},t_{\mathbf{x},2},\dots,t_{\mathbf{x},\text{N}_{\text{res}}})$ wrapped by the natural language question. The output is a free-form textual response representing the predicted answer.
Specifically, the prompt is constructed as: \texttt{[question tokens]} + \texttt{[sequence tokens]} + \texttt{[structure tokens]} + \texttt{[question tokens]}, and the output is a textual response represented as \texttt{[answer tokens]}.
In contrast, under the in-context learning setting, the model receives multiple protein QA examples as demonstrations before the final query.
Each demonstration includes both the prompt and its corresponding answer, mimicking how humans learn from examples.

\subsection{Model Overview}
\label{subsec_overview}
\begin{figure*}[t]
\centering
\includegraphics[width=1.0\textwidth]{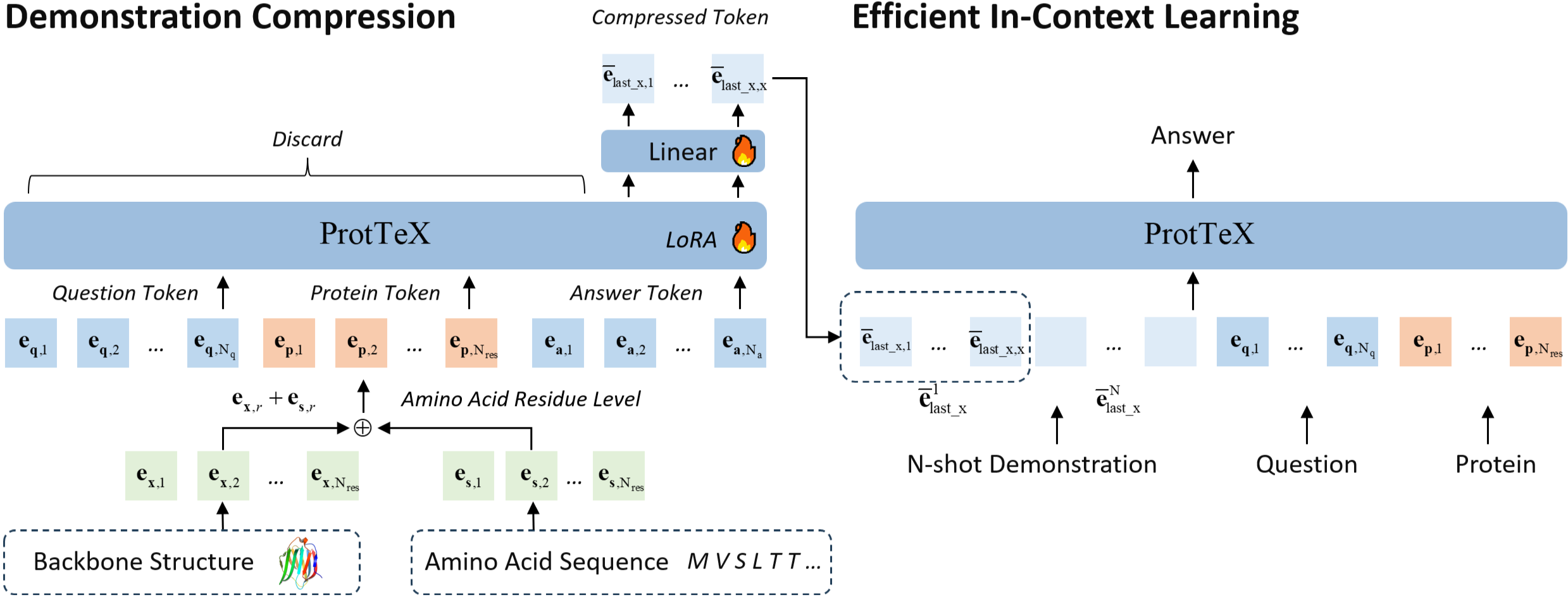}
\caption{Inference pipeline of our ProtTeX-CC.}
\label{fig_archi}
\end{figure*}
ProtTeX-CC is a two-stage compression framework designed to enable efficient in-context learning (ICL) for ProtTeX under limited context length.
As illustrated in Figure~\ref{fig_archi}, we propose a two-stage solution: 
\begin{itemize}
    \item A joint embedding compression stage that fuses the protein sequence and structure into unified residue-level representations, reducing token length by half while preserving semantic alignment.
    \item A self-compression stage that projects each full demonstration into a compact representation using only the last few linguistic tokens, reducing the average demonstration length from $751.41$ tokens to less than $16$ tokens.
\end{itemize}

During inference, the model consumes multiple compressed demonstrations along with the query in an in-context learning prompt format.
Compared to the original ProtTeX, our self-compression approach achieves a compression ratio of approximately $93.68\%$ in the total prompt length under the 16-shot setting.

\subsection{Joint Embedding Compression}
ProtTeX represents a protein with $\text{N}_{\text{res}}$ residues using two separate but aligned sequences of discrete tokens: the amino acid sequence tokens and the structure tokens.
However, the corresponding sequence and structure tokens, denoted by $(t_{\mathbf{s},r},t_{\mathbf{x},r})$, are separated by approximately $\text{N}_\text{res}$ tokens in the input stream, resulting in a substantial semantic and positional gap between the two modalities, which hinders the model’s ability to reason jointly over localized sequence–structure dependencies.
To mitigate this limitation, we introduce a joint embedding mechanism that integrates sequence and structure representations at the amino acid residue level as illustrated in Figure~\ref{fig_archi}.

\begin{align} 
\mathbf{e}_{\mathbf{p},r}=\mathbf{e}_{\mathbf{x},r} + \mathbf{e}_{\mathbf{s},r}
\label{eq_first_compress}
\end{align}
where $\mathbf{e}_{\mathbf{p}}=(\mathbf{e}_{\mathbf{p},1},\mathbf{e}_{\mathbf{p},2},\dots,\mathbf{e}_{\mathbf{p},\text{N}_{\text{res}}})$ denotes the joint representation of the protein, and $r$ is the amino acid residue index.
The sequence and structure embeddings are given by $\mathbf{e}_{\mathbf{x}}=f_{\phi,\text{embed}}(t_{\mathbf{x}})$ and $\mathbf{e}_{\mathbf{s}}=f_{\phi,\text{embed}}(t_{\mathbf{s}})$ respectively, where $f_{\phi,\text{embed}}$ is the embedding layer of ProtTeX.
We fine-tune this joint embedding representation on the PFUD dataset using LoRA~\cite{hu2021loralowrankadaptationlarge}, which facilitates efficient adaptation without full model updates and obtains $\text{ProtTeX}_{\text{joint embed}}$.
In our joint embedding compression stage, we reduce the input length and enhance the explicit correspondence between the amino acid sequence and structure.

\subsection{Information Aggregation of Last Few Tokens}
Motivated by observations in Table~\ref{tab_atten}, we construct efficient in-context demonstrations by retaining the final few tokens of the complete protein question-answer triplet.
These trailing tokens aggregate contextual information of the full demonstration through the forward propagation within the $\text{ProtTeX}_{\text{joint embed}}$.
To be specific, we feed the protein PDB file to the tokenization module of ProtTeX to obtain the separate sequence tokens $t_{\mathbf{s}}$ and structure tokens $t_{\mathbf{x}}$. Then we integrate sequence and structure embeddings at the amino acid residue level as described in Eq.~\ref{eq_first_compress}.
The aggregated tokens are obtained as follows:
\begin{align} 
\_,\mathbf{e}_{\text{last\_x}}=f_{\phi^{\prime},\text{forward}}(\mathbf{e}_{\mathbf{q}} \oplus \mathbf{e}_{\mathbf{p}} \oplus \mathbf{e}_{\mathbf{a}})
\label{eq_second_compress}
\end{align}
where $\mathbf{e}_{\mathbf{q}}$ is the question embeddings and $\mathbf{e}_{\mathbf{a}}$ is the answer embeddings. $\oplus$ means token-level concatenation.
$\mathbf{e}_{\text{last\_x}}$ denotes the last $\text{x}$ tokens of the full demonstration, where $\mathbf{e}_{\text{last\_x},i} \in \mathbb{R}^{d_f}, 1 \le i \le \text{x}$. And $d_f$ is the dimension of the hidden states of the backbone model. $f_{\phi^{\prime},\text{forward}}$ denotes the forward propagation of $\text{ProtTeX}_{\text{joint embed}}$. Considering $\mathbf{e}_{\text{last\_x}}$ lies in the output space of the inner protein LLM, we design a learnable projection layer parameterized by a linear layer serving as the adapter to convert each $\mathbf{e}_{\text{last\_x},i}$ into the protein LLM-acceptable token.
\begin{align} 
\bar{\mathbf{e}}_{\text{last\_x},i} = \text{Linear}(\mathbf{e}_{\text{last\_x},i})
\label{eq_linear}
\end{align}
where $\text{Linear}$ denotes the trainable projection layer of ProtTeX-CC.
The aggregation process of demonstrations is independent of each other, so we obtain the whole last $\text{x}$ embeddings $\mathbf{e}_{\text{last\_x}}$ of the demonstration set using $\text{ProtTeX}_{\text{joint embed}}$ for efficiency.

\subsection{Inference under ICL Settings}
To enable efficient few-shot inference, we construct $\text{N}$-shot compressed demonstrations by concatenating $\text{N}$ dependent last $\text{x}$ tokens as follows:
\begin{align} 
\mathbf{D}_{\text{N},\text{x}}=\bar{\mathbf{e}}^1_{\text{last\_x}}\oplus \bar{\mathbf{e}}^2_{\text{last\_x}} \oplus \dots \oplus \bar{\mathbf{e}}^{\text{N}}_{\text{last\_x}}
\label{eq_demo}
\end{align}
These compressed demonstrations serve as input context for autoregressive inference. Specifically, the frozen ProtTeX-CC model, parameterized by $\phi''$, receives the concatenation of $\mathbf{D}_{\text{N},\text{x}}$, the fully question embedding $\mathbf{e}_{\mathbf{q}}$, and the joint protein embeddings $\mathbf{e}_{\mathbf{p}}$ as contextual input to generate the answer tokens $\mathbf{e}_{\mathbf{a}}$ in an autoregressive manner:
\begin{align} 
\mathbf{e}_{\mathbf{a}}=\arg\max \mathrm{P}_{\phi''}(\mathbf{e}_{\mathbf{a},t}\mid \mathbf{D}_{\text{N},\text{x}};\mathbf{e}_{\mathbf{q}};\mathbf{e}_{\mathbf{p}};\mathbf{e}_{\mathbf{a},<t})
\label{eq_inference}
\end{align}

\subsection{Training Procedure}
We adopt a two-stage training strategy. In the first stage, we fine-tune $\text{ProtTeX}$ on the PFUD dataset using LoRA~\cite{hu2021loralowrankadaptationlarge} to fuse separate sequence and structure tokens into unified residue-level embeddings.
In the second stage, we freeze all model weights and train only a lightweight projection layer that maps compressed demonstrations into a format usable for few-shot inference.
Both stages are supervised with a standard language modeling loss:
\begin{align}
\text{loss}_1&=-\frac{1}{\lvert \mathbf{e}_{\mathbf{a}} \rvert} \sum_{t=0}^{\lvert \mathbf{e}_{\mathbf{a}} \rvert} \log \mathrm{P}_{\phi'}(\mathbf{e}_{\mathbf{a},t} \mid \mathbf{e}_{\mathbf{q}};\mathbf{e}_{\mathbf{p}};\mathbf{e}_{\mathbf{a},<t})\\
\text{loss}_2&=-\frac{1}{\lvert \mathbf{e}_{\mathbf{a}}\rvert} \sum_{t=0}^{\lvert \mathbf{e}_{\mathbf{a}}\rvert}\log \mathrm{P}_{\phi''}(\mathbf{e}_{\mathbf{a},t}\mid \mathbf{D}_{\text{N},\text{x}};\mathbf{e}_{\mathbf{q}};\mathbf{e}_{\mathbf{p}};\mathbf{e}_{\mathbf{a},<t})
\label{eq_loss}
\end{align}
where $\phi'$ and $\phi''$ denote the parameters of the joint embedding model and the self-compression model respectively;
$\mathbf{e}_{\mathbf{a}}$ is the answer sequence, $\mathbf{e}_{\mathbf{q}}$ and $\mathbf{e}_{\mathbf{p}}$ represent the encoded query and protein, and $\mathbf{D}_{\text{N},\text{x}}$ is the set of $\text{N}$ compressed demonstrations.

\subsection{Demonstration Retrieval Strategy}
\label{subsec_demo_selection}
To enable effective in-context learning, it is crucial to retrieve relevant demonstrations that are semantically aligned with the input query.
For demonstration selection, we utilize the compressed embeddings of all candidate examples.
Each compressed example $\mathbf{e}_{\text{last\_x}}$ is averaged by a pooled embedding vector $\mathbf{e}_{\text{candidate}}$ as follows:
\begin{align}
\mathbf{e}_{\text{candidate}} &= \text{Pool}\left( \mathbf{e}_{\text{last\_x}} \right)
\end{align}
We also use the frozen $\text{ProtTeX}_{\text{joint embed}}$ to encode both the query question $\mathbf{e}_{\mathbf{q},\text{query}}$ and query protein $\mathbf{e}_{\mathbf{p},\text{query}}$.
The query is similarly encoded into $\mathbf{e}_{\text{query}}$ using average pooling over question and protein token embeddings.
\begin{align}
\mathbf{e}_{\text{query}} &= \text{Pool}\left(f_{\phi^{\prime},\text{forward}}(\mathbf{e}_{\mathbf{q},\text{query}} \oplus \mathbf{e}_{\mathbf{p},\text{query}}) \right)
\end{align}
where $f_{\phi^{\prime},\text{forward}}$ denotes the forward propagation of $\text{ProtTeX}_{\text{joint embed}}$.
We then compute cosine similarity between the pooled representations:
\begin{align}
\text{sim}(\text{query}, \text{candidate}) = \frac{\mathbf{e}_{\text{query}}^\top \mathbf{e}_{\text{candidate}}}{\|\mathbf{e}_{\text{query}}\| \cdot \|\mathbf{e}_{\text{candidate}}\|}
\label{eq_sim}
\end{align}
In addition to embedding-based retrieval, we evaluate ProtTeX-CC under multiple demonstration selection strategies to comprehensively assess its in-context learning capabilities~\cite{gao2025uniiclefficientunifiedframework}. Specifically, we consider:
\begin{itemize}
  \item \textbf{Random}: Demonstrations are randomly sampled from the training set. This setting tests the model's ability to generalize from diverse, potentially unrelated examples.
  \item \textbf{BM25}: A sparse retrieval method commonly used in information retrieval. We use the BM25 algorithm to retrieve top-k demonstrations most relevant to the query based on lexical overlap.
\end{itemize}

\section{Experiment}

\begin{table*}[t]
\small
\centering
\setlength{\tabcolsep}{4pt}
\begin{tabular}{l|ccccc|ccccc}
\toprule
\multirow{2}{*}{Model} & \multicolumn{5}{c|}{PFUD (In-Domain)}& \multicolumn{5}{c}{UniProtQA (Out-of-Domain)}\\
\cmidrule(lr){2-11}
&\text{EMJI}&BLEU-2&ROUGE-1&ROUGE-2&ROUGE-L&\text{EMJI}&BLEU-2&ROUGE-1&ROUGE-2&ROUGE-L\\
\midrule
BioMedGPT-LM-10B$^{\dagger}$ &11.31&2.41&18.91&2.99&14.89&5.11&5.06&22.51&6.22&17.51\\
ProtT3$^{\dagger}$ &65.40&40.79&61.97&42.53&56.98&19.65&27.78&50.35&28.03&44.68\\
ProtTeX &71.73&41.54&63.46&43.17&57.89&19.64&26.66&49.45&25.94&42.34\\
ProtTeX (1-shot) &17.42&34.09&55.10&32.43&48.12&8.62&20.08&42.78&17.76&35.35\\
\midrule
$\text{ProtTeX}_{\text{joint embed}}$ &71.59&41.22&64.48&43.68&58.24&19.63&24.67&48.22&23.98&40.57\\
ProtTeX-CC (Random) &71.49&41.64&64.31&43.91&58.36&21.42&27.98&51.26&28.00&44.67\\
ProtTeX-CC (BM25) &71.92&\textbf{41.77}&63.60&43.54&58.41&21.70&28.45&51.67&28.50&45.02\\
ProtTeX-CC &\textbf{72.93}&41.28&\textbf{64.96}&\textbf{44.16}&\textbf{59.13}&\textbf{21.99}&\textbf{29.83}&\textbf{51.82}&\textbf{30.43}&\textbf{45.97}\\
\bottomrule
\end{tabular}
\caption{
Main results of multi-turn protein QA across PFUD and UniProtQA benchmarks.
All models (except $^{\dagger}$) are trained on PFUD and evaluated directly on UniProtQA without additional fine-tuning.
BioMedGPT-LM-10B$^{\dagger}$ and ProtT3$^{\dagger}$ were trained on both datasets.
ProtTeX (1-shot) represents the original ProtTeX applying naive in-context learning, which can fit at most one example within its context window.
$\text{ProtTeX}_{\text{joint embed}}$ denotes the zero-shot result of our joint embedding compression model.
ProtTeX-CC denotes the few-shot result of our self-compression variant ($\text{x}=16$, $\text{N}=16$).}
\label{tab_main_results}
\end{table*}

\subsection{Dataset}
\begin{figure}[t]
\centering
\includegraphics[width=0.28\textwidth]{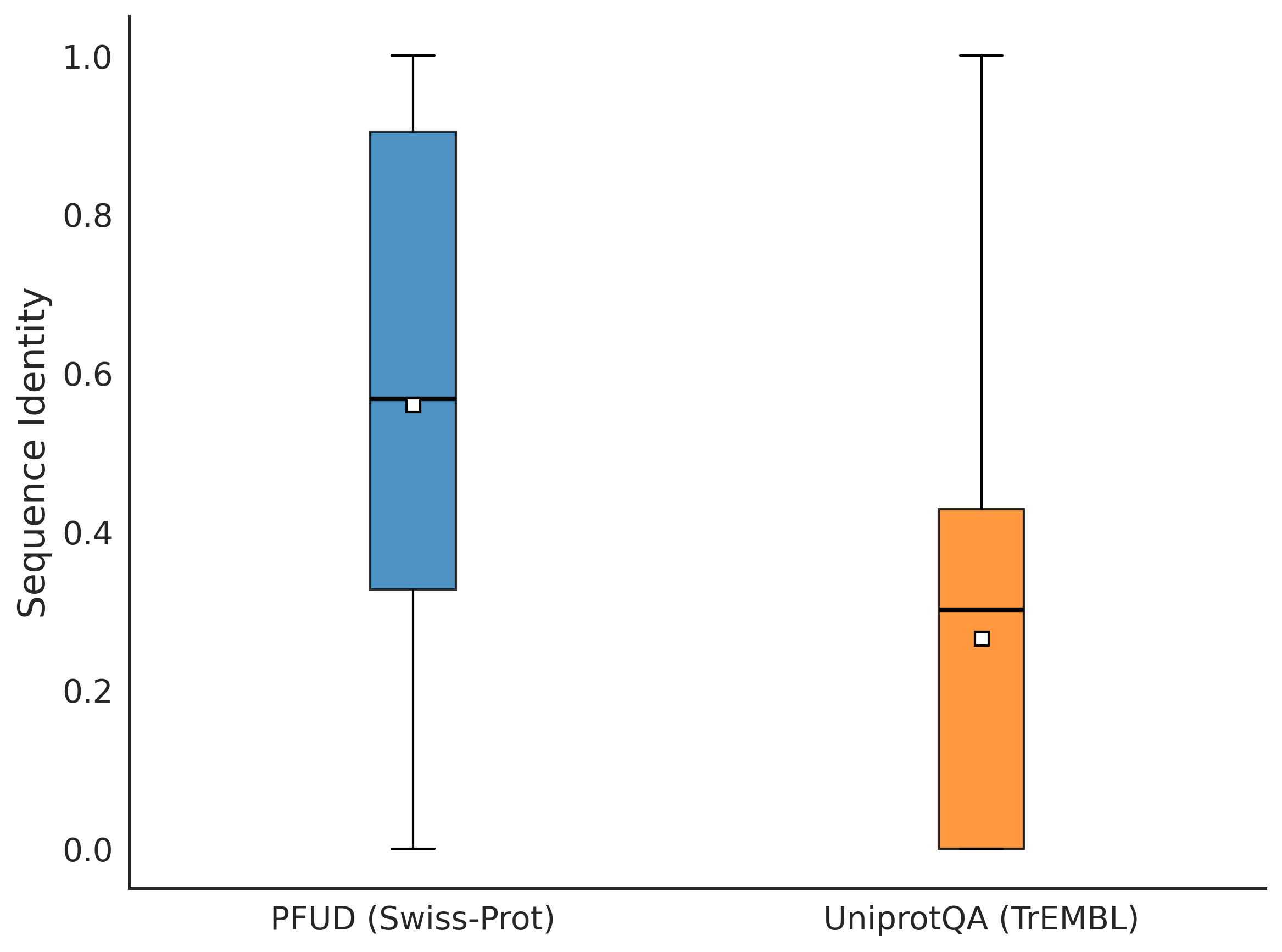}
\caption{Boxplot showing sequence identity distributions between the PFUD test and training sets, and between the UniProtQA test set and PFUD training set.}
\label{fig_dataset_distribution}
\end{figure}

To comprehensively evaluate the effectiveness and generalization of our method, we conduct experiments on two protein QA benchmarks. PFUD~\cite{ma2025prottex} serves as the primary training corpus for both the joint embedding compression and self-compression stages, while UniProtQA~\cite{luo2023biomedgptopenmultimodalgenerative} is used exclusively for evaluation to assess out-of-domain generalization.
As shown in Figure~\ref{fig_dataset_distribution}, the two datasets exhibit low sequence identity, highlighting their distributional divergence and the challenge of generalizing to unseen proteins.
For detailed dataset statistics and construction procedures, please refer to Appendix~\ref{appendix_dataset}.

\paragraph{PFUD}
PFUD consists of 426,928 protein QA pairs derived from Mol-Instructions~\cite{fang2024molinstructionslargescalebiomolecularinstruction}. To ensure high-quality annotations, we exclude entries from TrEMBL~\cite{10.1093/nar/27.1.49} and generate all examples using proteins exclusively from Swiss-Prot~\cite{10.1093/nar/27.1.49} via the same templating strategy. After tokenization with ProtTeX, the dataset contains approximately 320.8 million tokens.
The data is split into 90\% for training, 5\% for validation, and 5\% for testing.

\paragraph{UniProtQA} 
To assess generalization, we adopt the UniProtQA~\cite{luo2023biomedgptopenmultimodalgenerative} setup and construct an external test set of 2,500 TrEMBL~\cite{UniProt2025} proteins, each sharing less than 30\% average sequence similarity with PFUD training proteins (see Figure~\ref{fig_dataset_distribution}). The test set spans five question types—molecular function, subcellular location, biological process, protein caption, and multi-attribute—offering a diverse and challenging benchmark.

\subsection{Evaluation Metrics}
We evaluate model performance from two perspectives: general linguistic quality and domain-specific biological relevance.
For linguistic quality, we use BLEU and ROUGE to measure n-gram overlap with reference annotations.
For biological relevance, we adopt the Exact Match Jaccard Index (EMJI)~\cite{ma2025prottex}, which quantifies the overlap of functional Gene Ontology (GO) terms between predicted and reference outputs.
Formally, the EMJI for a test set of size $\text{N}_{\text{test}}$ is defined as:
\begin{align} 
\text{EMJI}=\frac{1}{\text{N}_{\text{test}}}\textstyle\sum_{i=1}^{\text{N}_{\text{test}}}\frac{\lvert\text{K}_{\text{label},i}\cap \text{K}_{\text{infer},i}\rvert}{\lvert\text{K}_{\text{label},i}\cup \text{K}_{\text{infer},i}\rvert}
\label{eq_emji}
\end{align}
where $\text{K}_{\text{label},i}$ and $\text{K}_{\text{infer},i}$ represents the set of biological keyworks extracted from the ground-truth label and model-generated answer for the test instance $i$, respectively.
To automatically extract keywords and perform exact matching, we leverage the DeepSeek-V3 model~\cite{deepseekai2025deepseekr1incentivizingreasoningcapability}, following the keyword extraction prompting protocol described in~\cite{ma2025prottex}.

\subsection{Baselines}
We compare our approach with several baselines, including the original ProtTeX~\cite{ma2025prottex}, and its 1-shot variant that applies naive ICL within its context window; $\text{ProtTeX}_{\text{joint embed}}$, our variant with residue-level joint embedding compression; and ProtTeX-CC, our full self-compression model leveraging the last few tokens per demonstration. We also include BioMedGPT-LM-10B~\cite{luo2023biomedgptopenmultimodalgenerative}, a domain-specific LLM pretrained on biomedical corpora including UniProtQA, and ProtT3~\cite{liu2024prott3proteintotextgenerationtextbased}, a Q-Former based multimodal protein LLM finetuned on Swiss-Prot~\cite{bairoch2000swiss}, ProteinKG25~\cite{zhang2022ontoprotein}, and PDB-QA~\cite{guo2023proteinchat}, which overlaps with UniProtQA.

\subsection{Setting}
For joint embedding learning, the model is trained on four GeForce RTX 3090 GPUs for one epoch using the AdamW optimizer with a learning rate of $2 \times 10^{-5}$ and a batch size of $4$.
For projection layer training in the self-compression stage, we retrieve $\text{N}$ in-context demonstrations for each query instance, where $\text{N} \in \{1, 2, 4, 8, 16\}$.
In addition, we investigate different levels of compression by varying the number of retained tokens per demonstration in the self-compression module, using the configuration of last $\text{x}$, where $\text{x} \in \{1, 2, 4, 8, 16\}$.
The demonstration set is the training set for both benchmarks.
To evaluate the few-shot in-context learning performance under varying demonstration counts, we experiment on the validation set and identify the optimal setting.
Furthermore, to ensure the reproducibility of our results, we employ greedy search for all protein understanding tasks in both benchmarks.

\subsection{Main Results}
As shown in Table~\ref{tab_main_results}, on the In-Domain PFUD benchmark, ProtTeX-CC achieves comparable performance to the original ProtTeX model.
As the demonstrations used for in-context learning are extracted from proteins encountered during training, the improvements on PFUD are relatively contained, reflecting the limited information introduced by the examples.
A pronounced performance drop is observed when applying naive in-context learning with the original ProtTeX, as adding demonstrations may introduce noise rather than provide useful context—likely due to the lack of training on interleaved multimodal inputs.
What's more, even with a 50\% reduction in protein length, $\text{ProtTeX}_{\text{joint embed}}$ performs on par with the original ProtTeX on the PFUD benchmark. This highlights the efficiency of our first-stage compression in maintaining representational fidelity.

In contrast, on the Out-of-Domain UniProtQA benchmark, which contains proteins from broader and more diverse sources, we observe substantial performance gains against the ProtTeX baseline when leveraging compressed in-context demonstrations.
Specifically, ProtTeX-CC achieves improvements of $11.26\%$ in EMJI, $11.89\%$ in BLEU-2, and an average ROUGE gain of $10.23\%$ (averaged over ROUGE-1, ROUGE-2, and ROUGE-L).
Notably, ProtTeX-CC variants with either random or similarity-based retrieval demonstrations consistently outperform the zero-shot ProtTeX baseline across all metrics, and even surpass the ProtT3 and BioMedGPT, which were finetuned on the UniProtQA or related datasets.
This highlights the robustness of our compression framework and its ability to transfer protein knowledge without explicit fine-tuning on the target domain.

\section{Analysis}
\begin{table*}[t]
\scriptsize
\centering
\begin{tabular}{p{6.8cm}|p{3.8cm}|p{2.4cm}|p{3cm}}
\Xhline{0.8pt}
\textbf{ProtTeX (Zero-shot)} & \textbf{ProtTeX-CC} & \textbf{Retrieved Example} & \textbf{Ground Truth} \\
\hline
The protein characterized by the structure and amino acid sequence demonstrates calcium channel regulator activity, calcium-dependent protein binding, ion binding, phospholipid binding, toxin activity and is implicated in the arachidonic acid secretion, defense response, hemostasis, negative regulation of T cell proliferation, phospholipid metabolic process. Its subcellular localization is primarily within the extracellular region.&Based on the given structure and amino acid sequence, the protein appears to have a primary function of \textit{\textbf{metal ion binding}}, \textit{\textbf{metalloendopeptidase activity}}. It \textit{\textbf{is likely involved in the proteolysis}}, and its subcellular localization is within the extracellular region.& The provided protein with given structure and sequence \textit{\textbf{is likely associated with the metal ion binding}}.&Upon analysis of the specified structure and amino acid sequence, it is evident that \textit{\textbf{the protein performs metalloendopeptidase activity}}, \textit{\textbf{zinc ion binding}}, \textit{\textbf{participating in the proteolysis}}.\\
\Xhline{0.8pt}
\end{tabular}
\caption{Case study on UniProtQA. Generated answers for UniProt protein A0A7J5X8E1 in response to the question: ``\textit{Could you analyze the protein corresponding to the protein structure and the amino acid sequence and offer insights on its function and the biological processes it might participate in?}". The retrieved example is the UniProt protein A0A1Y2DBZ1.}
\label{tab_case_study}
\end{table*}

\subsection{Analysis on Retrieval Strategies}
The results of different demonstration selection strategies show that our second-stage compressed representations are not only length compact but also semantically meaningful, enabling effective similarity-based retrieval for in-context learning.
Remarkably, despite the extreme compression length, these retrieved compressed embeddings yield performance gains of ProtTeX-CC on par with or even slightly outperforming BM25, a strong lexical matching baseline, across multiple evaluation metrics.
This demonstrates that our learned latent representations successfully capture high-level semantic similarities between protein examples, which traditional text-based methods like BM25 may miss due to the discrete and heterogeneous nature of protein sequences and structures.

\subsection{Analysis on Protein Tasks}
\begin{figure}[t]
\centering
\includegraphics[width=0.40\textwidth]{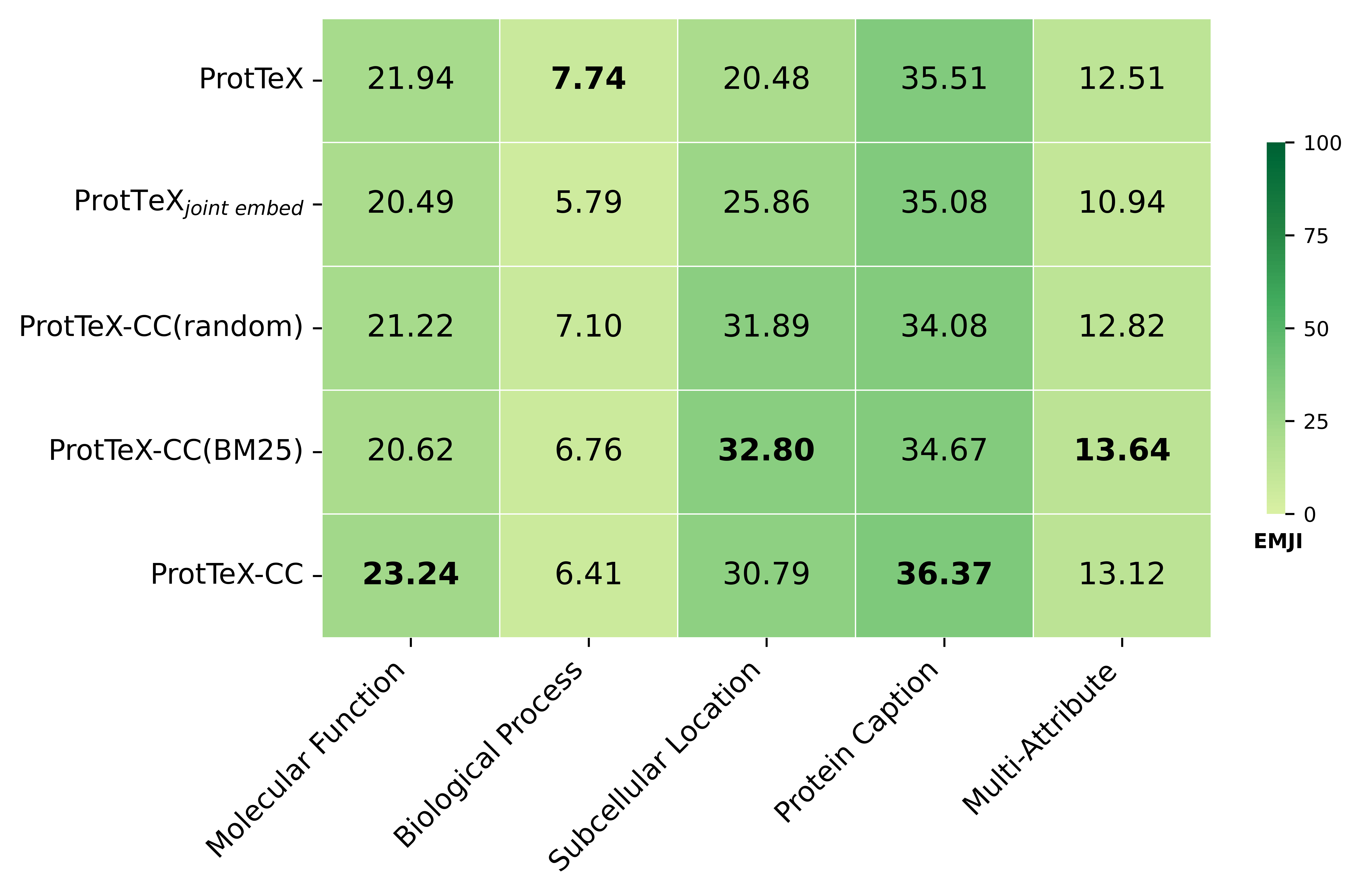}
\caption{EMJI across different protein understanding tasks in the UniProtQA benchmark.}
\label{fig_uniprot_emji}
\end{figure}

We evaluated the performance of ProtTeX-CC across five subtasks, as illustrated in Figure~\ref{fig_uniprot_emji}. ProtTeX-CC demonstrates improved accuracy in molecular function, subcellular localization, protein caption, and multi-attribute annotation, indicating enhanced capability in protein function understanding. By leveraging retrieval-based methods and in-context learning, ProtTeX-CC sheds light on the so-called ``dark regions" of proteins—those with limited or no prior functional annotations. This approach enables the model to effectively generalize from existing knowledge to previously uncharacterized proteins, thereby enhancing the performance of current models.

\subsection{Parameter Selection on Validation Set}
To select optimal self-compression parameters, we evaluate ProtTeX-CC on the UniProtQA validation set using embedding-based retrieval.
We systematically vary the number of retained tokens per demonstration ($\text{last\_x}$) and the number of in-context examples ($\text{N}$) to determine the configuration that balances compression and performance.
Table~\ref{tab_validation} presents ROUGE-L scores across different settings, along with retrieval Top-1 scores that reflect the quality of compressed demonstrations.
Compared to the zero-shot performance of ProtTeX and the retrieval Top-1 baseline, our approach yields consistent improvements across nearly all configurations.
What's more, results indicate that increasing the number of retained tokens generally enhances results, which demonstrates that the final linguistic tokens of demonstrations effectively capture essential task-relevant information, while our self-compression module enables efficient scaling to more in-context examples.

\begin{table}[t]
\scriptsize
\centering
\begin{tabular}{c|cccccc|c}
\Xhline{0.8pt}
\multirow{2}{*}{\text{x}}&\multicolumn{6}{c|}{$\#shots (\text{N})$}& \multirow{2}{*}{Retrieval Top-1}\\
\cline{2-7}
& 0&1&2&4&8&16& \\
\hline
1&\multirow{5}{*}{41.24}&41.22&42.05&41.89&42.84&44.02&30.19\\
2&&41.57&42.73&43.65&44.47&45.81&34.02\\
4&&43.02&44.49&45.85&47.43&48.20&32.54\\
8&&43.30&44.45&46.85&47.29&47.19&34.54\\
16&&\textbf{43.47}&\textbf{45.28}&\textbf{47.13}&\textbf{47.71}&\textbf{48.27}&29.23\\
\Xhline{0.8pt}
\end{tabular}
\caption{Validation set performance of ProtTeX-CC on UniProtQA under different configurations.
We report ROUGE-L (Retrieval Top-1) for each configuration.}
\label{tab_validation}
\end{table}

\subsection{Attention Analysis of the Instruction}
To better understand why compressed demonstrations remain effective under in-context learning, we conduct an attention-level analysis of $\text{ProtTeX}_{\text{joint embed}}$ on protein QA inputs. Table~\ref{tab_atten} reports the averaged attention scores assigned to different parts of the instruction, aggregated across all attention heads of the last layer of the backbone model.
Two key observations can be drawn from the results:
First, the model shows a clear bias toward the linguistic modality. Both the pre- and post-protein textual instructions receive substantially higher attention scores (0.72 and 1.38, respectively) than the protein tokens themselves (0.13). This suggests that ProtTeX relies more heavily on linguistic cues than on the raw protein representation.
Second, attention intensifies toward the latter part of the input, with the post-protein text receiving more weight. This reflects a positional bias of the built-in protein LLM.
These findings provide empirical support for our compression strategy in the second stage. Since the model primarily focuses on the concluding textual tokens within each demonstration, retaining only the final few tokens can still preserve key supervision signals—making these shortened tokens effective as efficient in-context demonstrations.
\begin{table}[t]
\centering
\small
\begin{tabular}{l|cc}
\Xhline{0.8pt}
Part of the Instruction & Avg. Score & Percentage \\
\hline
Text before the protein & 0.72 & 32.31\%\\
Joint protein representation& 0.14 & 6.20\%\\
Text after the protein & 1.38&61.49\%\\
\Xhline{0.8pt}
\end{tabular}
\caption{Averaged attention score from different parts of the instruction over multi-heads of $\text{ProtTeX}_{\text{joint embed}}$.}
\label{tab_atten}
\end{table}

\subsection{Case Study}
Table~\ref{tab_case_study} shows a representative case from UniProtQA (accession A0A7J5X8E1). The baseline ProtTeX model produces broad and irrelevant functions like ``calcium channel regulator activity" or ``toxin activity." In contrast, ProtTeX-CC accurately identifies \textit{metal ion binding} and \textit{metalloendopeptidase activity} by incorporating knowledge from the retrieved demonstration (accession A0A1Y2DBZ1) with similar functions. This highlights how our compressed examples preserve task-relevant semantics and enhance prediction precision on unseen proteins.

\section{Conclusion}
In this work, we address the inherent limitations of the discrete protein language model ProtTeX, in supporting multimodal in-context learning, particularly the inefficiency introduced by concatenated sequence-structure tokens.
We propose a two stage compression framework ProtTeX-CC that operates joint embedding compression to fuse sequence and structure representations, and self-compression to encode full demonstrations into compact latent tokens.
Without modifying the backbone model, ProtTeX-CC enables efficient in-context learning under constrained context lengths. Our approach achieves state-of-the-art performance on the PFUD benchmark and shows strong generalization on the unseen UniProtQA benchmark.

\bibliography{aaai2026}

\appendix

\section{ProToken}
ProtTeX tokenizes the protein backbone structure by following the method introduced in ProTokens~\cite{lin2025unifying,Lin2023.11.27.568722}, which focuses on tokenizing the metastable conformational structure $\mathbf{x}$ of proteins into discrete tokens via a reconstruction-based objective.
\begin{align} 
g_{\phi}(h_{\theta}(f_{\theta}(\mathbf{x})))\approx \mathbf{x}
\label{eq_ranking_loss}
\end{align}
where the encoder $f_{\theta}$ transforms a protein structure $\mathbf{x}$ into a $d$-dimensional latent representation $f_{\theta,r}(\mathbf{x}) \in \mathbb{R}^{d} $ for each residue $r$ ($1\le r \le \text{N}_{\text{res}}$). The tokenizer $h_{\theta}$ maintains a codebook $\{{\mathbf{c}_i}\}_{i=1}^{512}$ of $512$ code vectors, each $\mathbf{c}_i \in \mathbb{R}^{d}$. For each input vector $f_{\theta,r}(\mathbf{x})$ is assigned to the nearest code $\mathbf{c}_i$ via a nearest neighbor search, thus the ProToken for residue $r$ is defined as the code $\mathbf{z}_{\mathbf{x},r}= \arg\min_{\mathbf{c}_{i}\in \{\mathbf{c}_{i}\}}{\lVert f_{\theta,r}(\mathbf{x})-\mathbf{c}_i \rVert_{2}}^2$.
The decoder maps the ProToken sequence $\mathbf{z}_{\mathbf{x}}=(\mathbf{z}_{\mathbf{x},1},\mathbf{z}_{\mathbf{x},2},\dots,\mathbf{z}_{\mathbf{x},\text{N}_{\text{res}}})$ back to the input structure, such that $g_{\phi}(\mathbf{z}_{\mathbf{x}}) \approx \mathbf{x}$.
ProtTeX use the original amino acid sequence as the protein side-chain tokens.
Instead of using abbreviation letters or compositional encoders, ProtTeX extends the vocabulary by $512$ structural tokens and $20$ standard amino acid tokens.
This unified token space enables the model to process both sequence and structure in a consistent, discrete format.

\section{Dataset Details}
\label{appendix_dataset}
\paragraph{PFUD}
PFUD comprises 426,928 question-answer (QA) pairs derived from Mol-Instructions~\cite{fang2024molinstructionslargescalebiomolecularinstruction}. We excluded proteins annotated in TrEMBL~\cite{10.1093/nar/27.1.49} and expanded the dataset using proteins exclusively from Swiss-Prot~\cite{10.1093/nar/27.1.49}, following the same templating approach, ensuring that all protein entries in PFUD are sourced from Swiss-Prot. After tokenization with ProtTeX, the dataset contains approximately 320.8 million tokens. The data is split into 90\% for training, 5\% for validation, and 5\% for testing.
From the test set, we curated a benchmark consisting of 5,836 single-turn dialogues, covering six protein-related domains: molecular function (n = 1,127), subcellular location (n = 2,071), biological process (n = 459), domains or motifs (n = 886), protein captioning (n = 1,293), and multi-attribute questions (n = 974). Multi-attribute questions refer to those that involve multiple protein attributes within a single dialogue turn.
\paragraph{UniProtQA} 
TrEMBL contains unreviewed, automatically annotated entries in the UniProt~\cite{UniProt2025} database. To evaluate model generalization, we followed the approach of UniProtQA~\cite{luo2023biomedgptopenmultimodalgenerative} and curated an external test set of 2,500 proteins from TrEMBL that exhibit less than 30\% average sequence similarity to the proteins in the PFUD training set, as shown in Figure~\ref{fig_dataset_distribution}.
The corresponding questions were designed to mirror the format of those in PFUD, but with slightly modified prompts to assess the model's robustness to linguistic variation.
This external test set includes five question types: molecular function, subcellular location, biological process, protein caption, and multi-attribute questions, ensuring a comprehensive evaluation across diverse protein attributes.
This external test set includes five question types: molecular function (n = 500), subcellular location (n = 500), biological process (n = 500), protein captioning (n = 500), and multi-attribute questions (n = 500), ensuring a comprehensive evaluation across diverse protein attributes.

\section{Baseline Details}
\begin{itemize}
    \item ProtTeX: The baseline discrete protein large language model~\cite{ma2025prottex} that takes protein sequence and structure tokens as input.
    \item $\text{ProtTeX}_{\text{joint embed}}$: Our proposed joint embedding compression model that fuses sequence and structure information at the amino acid residue level.
    \item $\text{ProtTeX-CC}$: Our proposed self-compression variant that uses only the last $\text{x}$ tokens from each demonstration to fit more efficient examples within the context window.
    \item BioMedGPT-LM-10B: A domain-specific large language model introduced in~\cite{luo2023biomedgptopenmultimodalgenerative}, pretrained on biomedical corpora including UniProtQA. It encodes protein sequences using ESM-3B~\cite{lin2022language} and employs BioMedGPT-LM-7B as the decoder for response generation.
    \item ProtT3: A multimodal model~\cite{liu2024prott3proteintotextgenerationtextbased} that leverages ESM-3B to encode protein sequences, followed by a Q-Former to align protein embeddings with protein LLM. ProtT3 was finetuned on Swiss-Prot~\cite{bairoch2000swiss}, ProteinKG25~\cite{zhang2022ontoprotein}, and PDB-QA~\cite{guo2023proteinchat}.
\end{itemize}

\section{Limitations}
While ProtTeX-CC demonstrates strong performance and generalization under few-shot settings, several limitations remain. From our parameter analysis experiment, we observe that increasing the number of in-context examples ($\text{N}$) generally leads to better performance, and the trend has not yet converged at $\text{N}=16$. This suggests that scaling laws may also apply in the context of compressed in-context learning, and further performance gains could be achieved with higher $\text{N}$ values.
Future work can explore more efficient prompt packing strategies or architectural modifications to accommodate more examples. Additionally, a trade-off exists between the number of retained tokens per demonstration (i.e., $\text{last\_x}$) and overall compression ratio. Identifying an optimal $\text{last\_x}$ that balances compression efficiency with representational fidelity remains an open research direction.
\end{document}